\documentclass{article}

\usepackage[square,sort,comma,numbers]{natbib}
\usepackage[final]{MLWEB3_2022}

\usepackage{listings}

\usepackage{listings, xcolor}

\definecolor{verylightgray}{rgb}{.97,.97,.97}

\lstdefinelanguage{Solidity}{
	keywords=[1]{anonymous, assembly, assert, balance, break, call, callcode, case, catch, class, constant, continue, constructor, contract, debugger, default, delegatecall, delete, do, else, emit, event, experimental, export, external, false, finally, for, function, gas, if, implements, import, in, indexed, instanceof, interface, internal, is, length, library, log0, log1, log2, log3, log4, memory, modifier, new, payable, pragma, private, protected, public, pure, push, require, return, returns, revert, selfdestruct, send, solidity, storage, struct, suicide, super, switch, then, this, throw, transfer, true, try, typeof, using, value, view, while, with, addmod, ecrecover, keccak256, mulmod, ripemd160, sha256, sha3}, 
	keywordstyle=[1]\color{blue}\bfseries,
	keywords=[2]{address, bool, byte, bytes, bytes1, bytes2, bytes3, bytes4, bytes5, bytes6, bytes7, bytes8, bytes9, bytes10, bytes11, bytes12, bytes13, bytes14, bytes15, bytes16, bytes17, bytes18, bytes19, bytes20, bytes21, bytes22, bytes23, bytes24, bytes25, bytes26, bytes27, bytes28, bytes29, bytes30, bytes31, bytes32, enum, int, int8, int16, int24, int32, int40, int48, int56, int64, int72, int80, int88, int96, int104, int112, int120, int128, int136, int144, int152, int160, int168, int176, int184, int192, int200, int208, int216, int224, int232, int240, int248, int256, mapping, string, uint, uint8, uint16, uint24, uint32, uint40, uint48, uint56, uint64, uint72, uint80, uint88, uint96, uint104, uint112, uint120, uint128, uint136, uint144, uint152, uint160, uint168, uint176, uint184, uint192, uint200, uint208, uint216, uint224, uint232, uint240, uint248, uint256, var, void, ether, finney, szabo, wei, days, hours, minutes, seconds, weeks, years},	
	keywordstyle=[2]\color{teal}\bfseries,
	keywords=[3]{block, blockhash, coinbase, difficulty, gaslimit, number, timestamp, msg, data, gas, sender, sig, value, now, tx, gasprice, origin},	
	keywordstyle=[3]\color{violet}\bfseries,
	identifierstyle=\color{black},
	sensitive=false,
	comment=[l]{//},
	morecomment=[s]{/*}{*/},
	commentstyle=\color{gray}\ttfamily,
	stringstyle=\color{red}\ttfamily,
	morestring=[b]',
	morestring=[b]"
}

\AtBeginDocument{%
  \providecommand\BibTeX{{%
    \normalfont B\kern-0.5em{\scshape i\kern-0.25em b}\kern-0.8em\TeX}}}

\usepackage[utf8]{inputenc} 
\usepackage[T1]{fontenc}    
\usepackage{hyperref}       
\usepackage{url}            
\usepackage{booktabs}       
\usepackage{amsfonts}       
\usepackage{nicefrac}       
\usepackage{microtype}      
\usepackage{xcolor}         
\usepackage{graphicx}

\title{A Blockchain Protocol for Human-in-the-Loop AI}

\author{
  Nassim Dehouche\thanks{https://www.ndehouche.github.io} \\
  Mahidol University International College\\
  Mahidol University\\
  Salaya, Thailand 73170 \\
  \texttt{nassim.deh@mahidol.edu} \\
  \And
  Richard Blythman\\
  Algovera\\
  Dublin \\
  Ireland \\
  \texttt{richard@algovera.ai} 
}

\begin{document}

\maketitle

\begin{abstract}
 Intelligent human inputs are required both in the training and operation of AI systems, and within the governance of blockchain systems and decentralized autonomous organizations (DAOs). This paper presents a formal definition of Human Intelligence Primitives (HIPs), and describes the design and implementation of an Ethereum protocol for their on-chain collection, modeling, and integration in machine learning workflows. 
\end{abstract}

\section{Introduction and Related Work}

Modern Artificial Intelligence tends to focus on centralization, autonomy and competition with humans \cite{aifails}. However, the idea of augmenting human intelligence \cite{cybernetics} and "man-computer symbiosis" \cite{symbiosis} was prevalent in the early days of AI and cybernetics.

Human-in-the-loop (HITL) machine  learning \cite{HITL} is a promising development in this regard. Intelligent human inputs are often included in the machine learning workflow before training, in the form of data annotation. The HITL approach extends the scope of this integration to include human-machine interactions during training, e.g. through expert supervision \cite{HITL2}, and post-training, e.g. in safety audits and fine-tuning models \cite{audits}.

Software applications for crowdsourcing human intelligence tasks face challenges pertaining to the unfair compensation of labor \cite{labor}, fraud \cite{fraud}, censorship \cite{censorship}, and the difficulty of vetting credentials \cite{credentials}. The latter is typified by protocols geared towards centralized crowd-labor platforms, such as Turkit \cite{turkit}. For example, human input is taken from an indistinct mass of crowd workers on Amazon's Mechanical Turk platform, without the ability to require a certain level of expertise or credentials from respondents. Turkit \cite{turkit} introduced the useful concept of scripting  human intelligence tasks within traditional web applications, and was designed with issues related to high-cost and high-latency steps involving humans in mind. This required engineering a \textit{crash-rerun} approach to avoid re-executing expensive steps. 

Decentralized software deployed on public, permissionless blockchains offers natural opportunities to tackle the aforementioned challenges.
Any write instruction in a smart contract is an atomic transaction that is immutably stored on the blockchain, and transparently accessible to any client application. Moreover, in addition to trustless, uncensorable payment processing, blockchain software can offer participants ownership in the system they partake in. Lastly, the emergence of standards for identity management, such as the non-fungible token standard, allow for sophisticated access control and have propelled the emergence of domain-expert decentralized autonomous organizations (DAOs).

DAOs are sometimes imagined as being governed by autonomous algorithms, with humans at the margins. However, there is an increasing push towards a future of collective intelligence that promotes harmony between humans and algorithms by optimizing for the autonomy of individuals \cite{autonomy}. We believe that protocols that facilitate crowdsourcing of human intelligence and preferences are a key component of this. This has applications in collection and annotation of training data, and AI safety.

In the following, we describe a protocol for Ethereum Virtual Machine (EVM)-compatible blockchains that allow for the on-chain modeling of human intelligence tasks and their integration in machine learning workflows.

\section{Human Intelligence Primitives}

A Human Intelligence Primitive (HIP) is a procedure for the collection and representation of preferences structures on a finite set of $n$ potential alternatives (i.e. comparable objects or actions) $A = \{a_0, \dots, a_{n-1}\}$. Preferences can be of one of the four following types, based on \cite{Roy}:

\begin{itemize}
    \item A choice $(P_1)$, that is a subset $A' \subseteq A$ of potential alternatives, typically a singleton set, containing the preferred alternative(s).
    \item A ranking $(P_2)$, that is a total preorder on $A$, ordering alternatives by decreasing preference, with possible \textit{ex-aequo}. A particular case of ranking is preferential voting \cite{preferential}, in which this preference structure is a total order on $A$.
    \item A sorting $(P_3)$, that is the assignment of each alternative in $A$ into pre-defined classes $C=\{c_0, \dots, c_{k-1}\}$, ordered by decreasing preference. A particular case of sorting is score voting \cite{score} on a discrete scale.
    \item  A classification $(P_4)$, that is the assignment of each alternative in $A$ into pre-defined, unordered classes $C=\{c_0, \dots, c_{k-1}\}$.

\end{itemize}
A HIP can thus be abstractly characterized by a triplet $(t, n, k)$, where $t \in \{P_1, P_2, P_3, P_4\}$ is the type of preferences sought, $n \in \{2, 3, \dots, +\infty\}$ the number of alternatives considered, and $m \in \{1, 2, \dots, +\infty\}$ the number of classes (equal to $1$ for a choice or ranking primitive).

\section{Smart Contract Architecture}
We propose a smart contract implementing HIPs to incentivize and coordinate  collective intelligence by humans within DAOs. HIPs can be initiated by Externally-Owned Account (EOA) on Ethereum, or contracts, through a CALL or DELEGATECALL operation. Conversely, the smart contract can communicate synchronous events to off-chain clients (e.g. the submission of a response to a HIP), and its output can be read asynchronously by these programs.

We consider two categories of users of the smart contract; \textit{proposers} and \textit{respondents}.
A HIP is recorded in a HIP object, the creation of which is initiated by a \textit{proposer} address, through a function \texttt{submitHIP()}. In this function, a proposer submits a triplet $(t, n, k)$, and pays a fee that depends on the type of primitive $t$. The type $t$ is encoded by an enumerable \texttt{types \{CHOICE, RANKING, SORTING, CLASSIFICATION}\}, while the number of alternatives $n$ and the number of classes $k$ are stored in unsigned integers. 

Additionally, the proposer specifies a duration, encoded as an unsigned integer number of seconds, for taking responses. This duration is relative to the creation date of the HIP, recorded as the timestamp of the valid block enacting its creation on the blockchain. Lastly, the HIP object records the number of responses (individual preference structures compatible with the HIP type) recorded so far, in an unsigned integer variable \texttt{numResponses}.

An individual response to a HIP is submitted by a \textit{respondent} address, through a function \texttt{submitResponse()}, specifying the address of the proposer, and the index of the HIP being responded to, among their proposed HIPs.
Respondents' access is gated by a non-fungible token (NFT), vetting their credentials, and giving them \textit{read} access to the corresponding off-chain semantic data, and \textit{write} access to record a response to a HIP in the contract. 
Moreover, before recording a response, we verify that the respondent has not already voted, and that the submitted response is compatible with the HIP type $t$. 
 An individual response that passes these checks is recorded in a \texttt{Response} structure, containing the address of the respondent and an array of unsigned integers, representing the content of the response.

Given the strengths of the blockchain (trustless access control and payment processing), and its weaknesses (inability to store secrets and high cost of computation), we have made the following key architecture choices:
\begin{itemize}
\item Since data are transparently stored on the blockchain, HIP objects are recorded in the abstract form of a triplet $(t, n, k)$, and linked with semantic data (i.e. descriptions for alternatives and classes) that are stored off-chain.
\item In order to incentivize responses, and discourage their concentration in a few HIPs, the reward of each respondent is the fee paid by the proposer divided by the number of responses, by the end of a HIP's duration.
\item Once recorded in the contract, responses can be eventually accessed by off-chain clients for computationally complex processing and aggregation.

\end{itemize}

\begin{figure}[h]

\centering
\includegraphics[width=\textwidth]{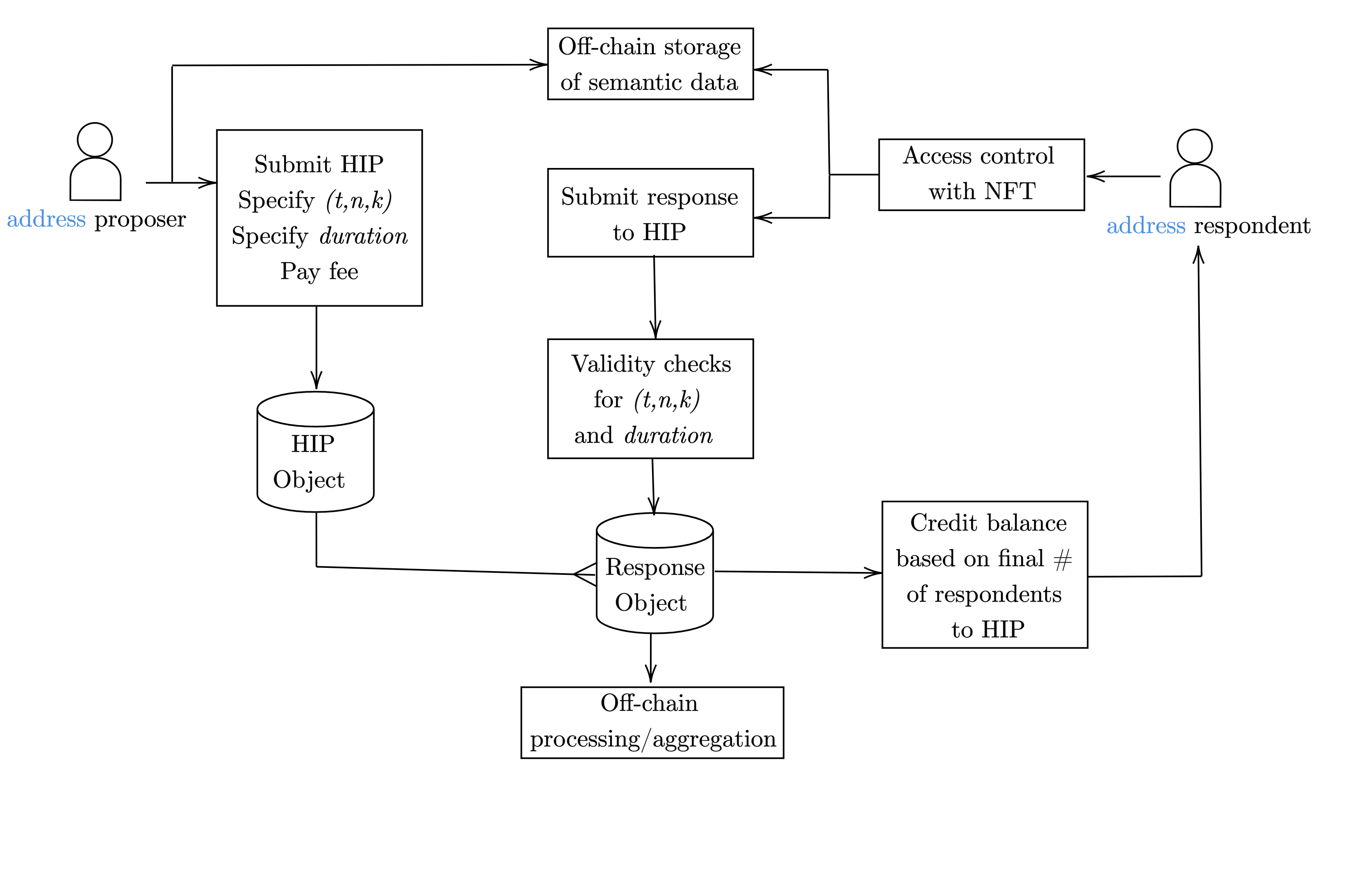}
\caption{\label{flow} Architecture of the proposed protocol}
\end{figure}

The proposed architecture is summarized in the process diagram in Figure \ref{flow}.

\section{Main Data Structures}
\begin{figure}[h]

\centering
\includegraphics[width=\textwidth]{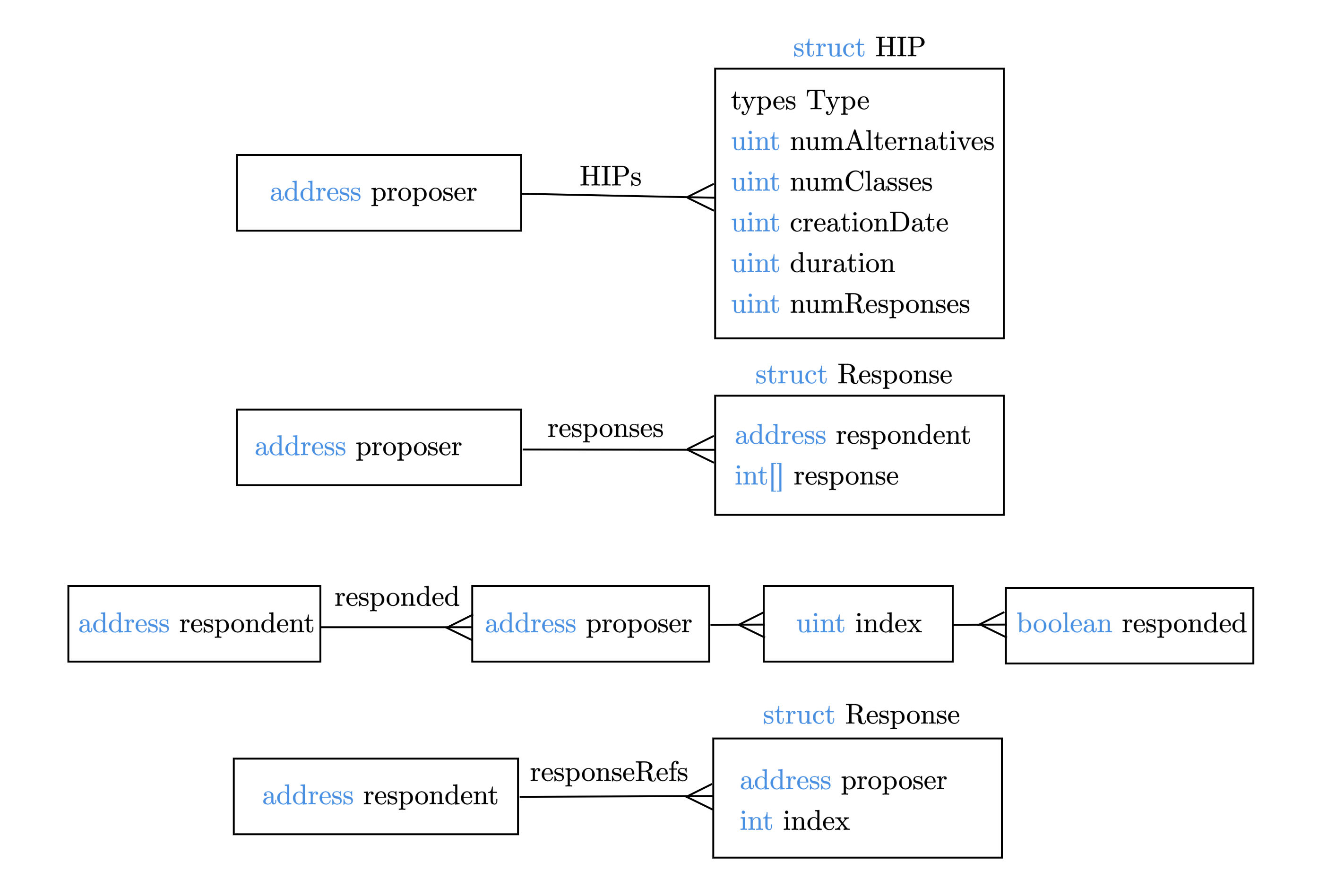}
\caption{\label{data} Main data structures}
\end{figure}

The implementation of the protocol is subject to four indexing requirements:
\begin{itemize}
    \item Reading/Writing HIPs requires mapping proposers with the HIPs they have created. 
    This is implemented as a \texttt{mapping(address => HIP[])}, named  \texttt{HIPs}, whose key is a proposer address, and value is an array of HIPs submitted by this address. 
    \item Reading/Writing Responses requires mapping HIPs with the responses they have received.  This is implemented as a double mapping, \texttt{mapping(address => mapping(uint => Response[]))}, named  \texttt{responses}, indexed by a proposer address and an integer index for a HIP, and whose value is an array of responses submitted for it. 
    \item Ensuring single responses requires mapping respondents and HIPs, with a boolean indicating whether the former has submitted a response to the latter. This is implemented as a triple mapping, \textit{mapping(address => mapping(address => mapping (uint =>bool)))}, named  \texttt{responded}, indexed by a respondent address, a proposer address and an integer index for a HIP, and whose value is a boolean indicating the existence of a response. 
    \item Payment processing requires mapping respondents with the proposers and indices of the HIPs they have responded to. This is implemented as a mapping, \texttt{mapping(address => ResponseRef[])}, named  \texttt{responseRefs}, indexed by a respondent address, and whose value is an array of objects of type \texttt{responseRef}, containing the address of a proposer and the index of a HIP.  
\end{itemize}
These four mappings are illustrated in Figure \ref{data}.
\subsection{Response Verification}
Before a response, submitted in the form of an array $R$ of unsigned integers, is recorded in the contract, we must verify that it is valid for a given HIP, defined by a triplet $(t, n, k)$.
\begin{itemize}
\item If $t$ indicates a choice primitive, the validity requirements are that $length(R)==1$ (i.e. we only allow singleton choices\footnote{This requirement could be changed to an inequality to allow for larger choice subsets.}) and $R[0]<n$ (i.e. the submitted choice corresponds to the index of a possible alternative).
\item If $t$ indicates a ranking primitive, the validity requirements are that $length(R)==n$ and $R$ contains unique digits between $0$ and $n-1$. This latter requirement is verified by a function $uniqueDigits()$ in $O(n)$, which uses a local boolean array variable of size $n$. Depending on the preferred storage-computation trade-offs, an alternative would be to verify the uniqueness of the digits of $R$ in $O(n^2)$, without the use of a local array. This is the most computationally-intensive potential operation in the proposed protocol.
\item If $t$ indicates a sorting or classification primitive, the validity requirements are that $length(R)==n$ and $R$ only contains digits between $0$ and $k-1$.

\end{itemize}

\subsection{Payment Processing}
Compensating respondents to a HIP proportionally to its total number of respondents poses challenges for payment processing. It notably not allow for a real-time incrementation of a respondent's balance. This is due to the fact that any computation on the EVM must be initiated by an EOA, and it does not allow for automated code execution. The solution we propose is to compute this balance once a respondent requests a payment, using the \texttt{requestPayment()} function, so that they can bear the gas cost of this computation.

\section{Example Applications}
A wide range of tasks requiring human intelligence can be expressed as HIPs, for example within the training and operation of AI systems and the governance of blockchain systems and DAOs. Surveys can be modeled as an instance of $P_1$ \cite{surveys}, the collection of training data for machine-learned ranking (MLR) 
as $P_2$ \cite{mlr}, independent AI safety audits as $P_3$ \cite{audits}, or data annotation as $P_4$ \cite{annotation}. In these examples, HIPs are used with a descriptive intent and a collection of individual preferences is their intended output. Moreover, when combined with a systematic aggregation procedure for individual preferences, HIPs can serve as primitives in processes such as plurality voting or approval voting as instances of $P_1$ \cite{plurality}, preferential voting
as $P_2$ \cite{preferential}, score voting as $P_3$ \cite{score}, or rule-based classification as $P_4$ \cite{classification}.

Following is an example, in pseudo-JavaScript, for a data annotation use case. The prefix "\texttt{contract.}" indicates a call to a function of the contract by an EOA or a web client, e.g. using the \textit{web3.js} library \cite{web3js}.
\begin{itemize}
\item Proposer calls a classification HIP with \texttt{await contract.methods.submitHIP(CLASSIFICATION, n,      2, duration).send(\{from:accounts[0], value:fee\});}
\item After a delay corresponding to the value of the argument \texttt{duration}, proposer collects responses with \texttt{response=await contract.methods.getResponse(proposer,index, i).call();}
\item Proposer aggregates responses off-chain using e.g. the majority rule.
\end{itemize}

\section{Conclusion and Perspectives}
This paper described the design and implementation of an Ethereum protocol to to incentivize and coordinate collective intelligence by humans on-chain. Experiments using the proposed protocol will be conducted in the Algovera community, a DAO for data scientists, in order to identify new use cases and optimize gas usage for typical real-world applications. 

The detailed source code of the proposed implementation can be found in the Appendix of this paper.

\section*{Acknowledgements}

This publication has emanated from research conducted with the partial financial support of Algovera Grants under grant number 22/AG/R1/6.
The first author is grateful to the members of Algovera DAO for fruitful discussions.

{}

\appendix

\nonumber\section{Appendix: Source Code of the Proposed Protocol}
\begin{lstlisting}[language=Solidity, numbers=none] 

// SPDX-License-Identifier: CC BY 4.0
pragma solidity ^0.8.12;
/**
* @title Human-augemented Intelligence contract
* @author Nassim Dehouche
*/
import "@openzeppelin/contracts/interfaces/IERC721.sol";
contract HaAI {
address owner;
address tokenContract ;
// HIP types
enum types{ CHOICE, RANKING, SORTING, CLASSIFICATION}
uint numProposers;  
address[] proposers;
uint[] fees; 

  constructor(){
    owner = msg.sender;   
  }

  /**
  @param _tokenContract is the address of the ERC-721 contract to vet
  voters. We assume one address, one NFT, one vote.  
  Use 0xF5b2B5b042B253323cB96121ABad487C95d287ea on Kovan
  */
function initialize (address _tokenContract, uint[] calldata _fees )
public{
    require(msg.sender == owner);
    tokenContract = _tokenContract;
    fees=_fees;
}

// The HIP structure
struct HIP{ 
  types HIPType;
  uint numAlternatives;
  uint numClasses;
  uint creationDate;
  uint duration;
  uint numResponses;
  }

// Mapping proposers with an array of their proposed HIPs
mapping(address => HIP[]) public HIPs; 

// The Response struct for the content of the response.
struct Response{ 
  address respondent;
  uint[] response;
  }


// The Response reference struct for payment.
struct ResponseRef{ 
  address proposer;
  uint index;
  }

// Responses. The first key is the proposer address
mapping(address => mapping(uint => Response[])) internal responses;

// The Response boolean. The first key is the respondent address
mapping(address => mapping(address => mapping (uint =>bool))) public responded;

// The Response reference for payment. Mapping respondent with the HIPs they responded to.
mapping(address => ResponseRef[]) public responseRefs;

modifier onlyIfPaidEnough(types _HIPType) {
    require(msg.value==fees[uint(_HIPType)], "User did not pay the right fee for this HIP type.");
    _;
}

modifier onlyIfHoldsNFT(address _voter) {
    require(IERC721(tokenContract).balanceOf(_voter) > 0, "User does not hold the right NFT.");
    _;
} 

modifier onlyIfHasNotResponded(address _proposer, uint _id) {
    require(responded[msg.sender][_proposer][_id]==false, "User has already responded.");
    _;
} 

modifier onlyIfStillOpen(address _proposer, uint _id) {
    require(block.timestamp<=HIPs[_proposer][_id].creationDate+HIPs[_proposer][_id].duration, "This HIP is no longer open for responses.");
    _;
} 

function submitHIP
(  
  types _HIPType,
  uint _numAlternatives,
  uint _numClasses,
  uint _duration) 
public 
payable
onlyIfPaidEnough(_HIPType)

returns(uint _id)
{
bool condition;
if (_numAlternatives>=2){
   condition=true;
   if (_HIPType==types.SORTING || _HIPType==types.CLASSIFICATION){
    condition=_numClasses>=2;

    }
}
    
if(!condition) { revert('Trivial or invalid HIP'); }

_id= HIPs[msg.sender].length;
if (_id==0){
numProposers++;
proposers.push(msg.sender);    
}
HIPs[msg.sender].push();
HIPs[msg.sender][_id].HIPType = _HIPType;
HIPs[msg.sender][_id].numAlternatives = _numAlternatives;
HIPs[msg.sender][_id].numClasses = _numClasses;
HIPs[msg.sender][_id].creationDate = block.timestamp;
HIPs[msg.sender][_id].duration = _duration;
return _id;
}

function rightDigits (uint[] calldata _response, uint _number)
internal 
pure
returns(bool _right)
{
uint i;
_right=true;
while (i<_response.length){
 if (_response[i]>=_number){
   return false;
 } 
 unchecked{i++;}
}
return _right;
}

function uniqueDigits (uint[] calldata _response, uint _number)
internal 
pure

returns(bool _unique)
{
bool[] memory visited; 
uint i;
_unique=true;
while (i<_response.length){
 if (_response[i]>=_number || visited[_response[i]]==true){
   return false;
 }
 else{
   visited[_response[i]]=true;
 } 
 unchecked{i++;}
}
return _unique;
}



function submitResponse
(  
  address _proposer,
  uint _id,
  uint[] calldata _response) 
public 
onlyIfHoldsNFT(msg.sender)
onlyIfHasNotResponded(_proposer, _id)
onlyIfStillOpen(_proposer, _id)
returns(uint _number)
{
bool condition;

    if (HIPs[_proposer][_id].HIPType==types.CHOICE){
   condition=_response.length==1 && _response[0]<HIPs[_proposer][_id].numAlternatives;
   }
    else if (HIPs[_proposer][_id].HIPType==types.RANKING){
    condition=_response.length==HIPs[_proposer][_id].numAlternatives && uniqueDigits(_response, _response.length);
   
    }
    else if (HIPs[_proposer][_id].HIPType==types.SORTING || HIPs[_proposer][_id].HIPType==types.CLASSIFICATION){
    condition=_response.length==HIPs[_proposer][_id].numAlternatives && rightDigits(_response, HIPs[_proposer][_id].numClasses);

    }
  
    
if(!condition) { revert('Invalid response'); }
   
_number=responses[_proposer][_id].length+1;
HIPs[_proposer][_id].numResponses=_number;
responses[_proposer][_id].push();
responses[_proposer][_id][_number-1].respondent=msg.sender;
 for(uint i = 0; i < _response.length; ) {
responses[_proposer][_id][_number-1].response.push(_response[i]);
unchecked{i++;}
}
ResponseRef memory r;
        r.proposer = _proposer;
        r.index = _id;
        
responseRefs[msg.sender].push(r);
responded[msg.sender][_proposer][_id]=true;
return _number;
}

// Respondents payment function 
    function requestPayment() public 
    {
    uint _balance;
    uint _id;
    address _proposer;
    for (uint i=0;i<responseRefs[msg.sender].length;){
    _proposer=responseRefs[msg.sender][i].proposer;
    _id=responseRefs[msg.sender][i].index; 
    if (_proposer!=address(0) && block.timestamp>HIPs[_proposer][_id].creationDate+HIPs[_proposer][_id].duration)
    {
    responseRefs[msg.sender][i].proposer=address(0);        
    _balance+=fees[uint8(HIPs[_proposer][_id].HIPType)]/HIPs[_proposer][_id].numResponses;
    unchecked{i++;}
    }
    }
      (bool sent, ) = msg.sender.call{value: _balance}("");
        require(sent, "Failed to send Ether");
   
   }

 function getNumProposers() public view returns(uint _numProposers){
     return numProposers;
   }

 function getFee(uint i) public view returns(uint _fee){
     return fees[i];
   }
function getProposer(uint i) public view returns(address _proposer){
     return proposers[i];
   }  

function getHIPCount(address _proposer) public view returns(uint _count){
     return HIPs[_proposer].length;
   }

function getResponse(address _proposer, uint _indexHIP, uint _indexResponse) public view returns(uint[] memory _response){
     return responses[_proposer][_indexHIP][_indexResponse].response;
   }

function getBalance() public view returns(uint _balance){
   
    uint _id;
    address _proposer;
    for (uint i=0;i<responseRefs[msg.sender].length;){
    _proposer=responseRefs[msg.sender][i].proposer;
    _id=responseRefs[msg.sender][i].index; 
    if (_proposer!=address(0) && block.timestamp>HIPs[_proposer][_id].creationDate+HIPs[_proposer][_id].duration)
    {    
    _balance+=fees[uint8(HIPs[_proposer][_id].HIPType)]/HIPs[_proposer][_id].numResponses;
    unchecked{i++;}
    }
    }
    return _balance;
   }

}
\end{lstlisting}

\end{document}